\documentclass[conference,a4paper]{APSIPA2021}
\usepackage{amsmath}
\usepackage{graphicx}
\usepackage{multirow}
\usepackage{threeparttable}
\usepackage[backend=biber,style=ieee,]{biblatex}
\usepackage{amssymb}
\usepackage[linesnumbered,algoruled, vlined]{algorithm2e}
\usepackage{geometry}
\usepackage{fancyhdr}

\fancypagestyle{firststyle}{
  \fancyhf{}
  \fancyhead[C]{2023 Asia Pacific Signal and Information Processing Association Annual Summit and Conference (APSIPA ASC)}
}

\begin{document}

\title{IPFed: Identity protected federated learning \\for user authentication}

\author{
\authorblockN{
Yosuke Kaga\authorrefmark{1}, Yusei Suzuki\authorrefmark{1}, Kenta Takahashi\authorrefmark{1}
}

\authorblockA{
\authorrefmark{1}
Hitachi, Ltd., Japan}
}

\begin{minipage}[t]{160mm}

\section*{IEEE Copyright Notice}
\large
Copyright 2023 IEEE. Personal use of this material is permitted. Permission from IEEE must be obtained for all other uses, in any current or future media, including reprinting/republishing this material for advertising or promotional purposes, creating new collective works, for resale or redistribution to servers or lists, or reuse of any copyrighted component of this work in other works.\\

Accepted by 2023 Asia Pacific Signal and Information Processing Association Annual Summit and Conference (APSIPA ASC)
\end{minipage}
\newpage

\maketitle
\thispagestyle{firststyle}
\pagestyle{fancy}

\begin{abstract}
With the development of laws and regulations related to privacy preservation, it has become difficult to collect personal data to perform machine learning.
In this context, federated learning, which is distributed learning without sharing personal data, has been proposed.
In this paper, we focus on federated learning for user authentication.
We show that it is difficult to achieve both privacy preservation and high accuracy with existing methods.
To address these challenges, we propose IPFed which is privacy-preserving federated learning using random projection for class embedding.
Furthermore, we prove that IPFed is capable of learning equivalent to the state-of-the-art method.
Experiments on face image datasets show that IPFed can protect the privacy of personal data while maintaining the accuracy of the state-of-the-art method.

\end{abstract}

\section{Introduction}
\label{sec:introduction}
User authentication, such as face recognition, has recently achieved dramatic improvements in accuracy through the application of deep learning, and one of the reasons for this is that large numbers of images have been collected through web crawling and used as training data \cite{Raji21}.
However, in recent years, GDPR \cite{Voigt17} and other privacy-related regulations have made it more difficult to collect personal data.
In order to continue to improve the accuracy of user authentication, Privacy-Preserving Machine Learning (PPML) \cite{Rubaie19} is needed, which performs machine learning without directly collecting personal data.
One of the most popular methods in PPML is federated learning \cite{mcmahan17}, which can perform privacy-preserving distributed learning using personal data on the client, and it has attracted attention.
Federated learning has been applied to user authentication \cite{Yu20,Aggarwal21,Hosseini20,Hosseini21}.
However, these methods face the challenges of privacy of training data \cite{Yu20,Aggarwal21} and model accuracy \cite{Hosseini20,Hosseini21}.
We propose Identity Protected Federated Learning (IPFed) as a solution to this problem.
The main contributions of our work are as follows.
\begin{itemize}
\item We develop IPFed, which is a method to perform federated learning for user authentication while preserving the privacy of personal data using random projection for class embeddings.
\item We prove mathematically that IPFed can perform equivalent learning to the state-of-the-art method while preserving the privacy of the training data.
\item Experiments on face image datasets show that the proposed method can learn models with equivalent accuracy to the state-of-the-art method.
\end{itemize}

\section{Related works}
\label{sec:related}
\begin{figure}[t]
\begin{center}
\includegraphics[clip,width=8.0cm]{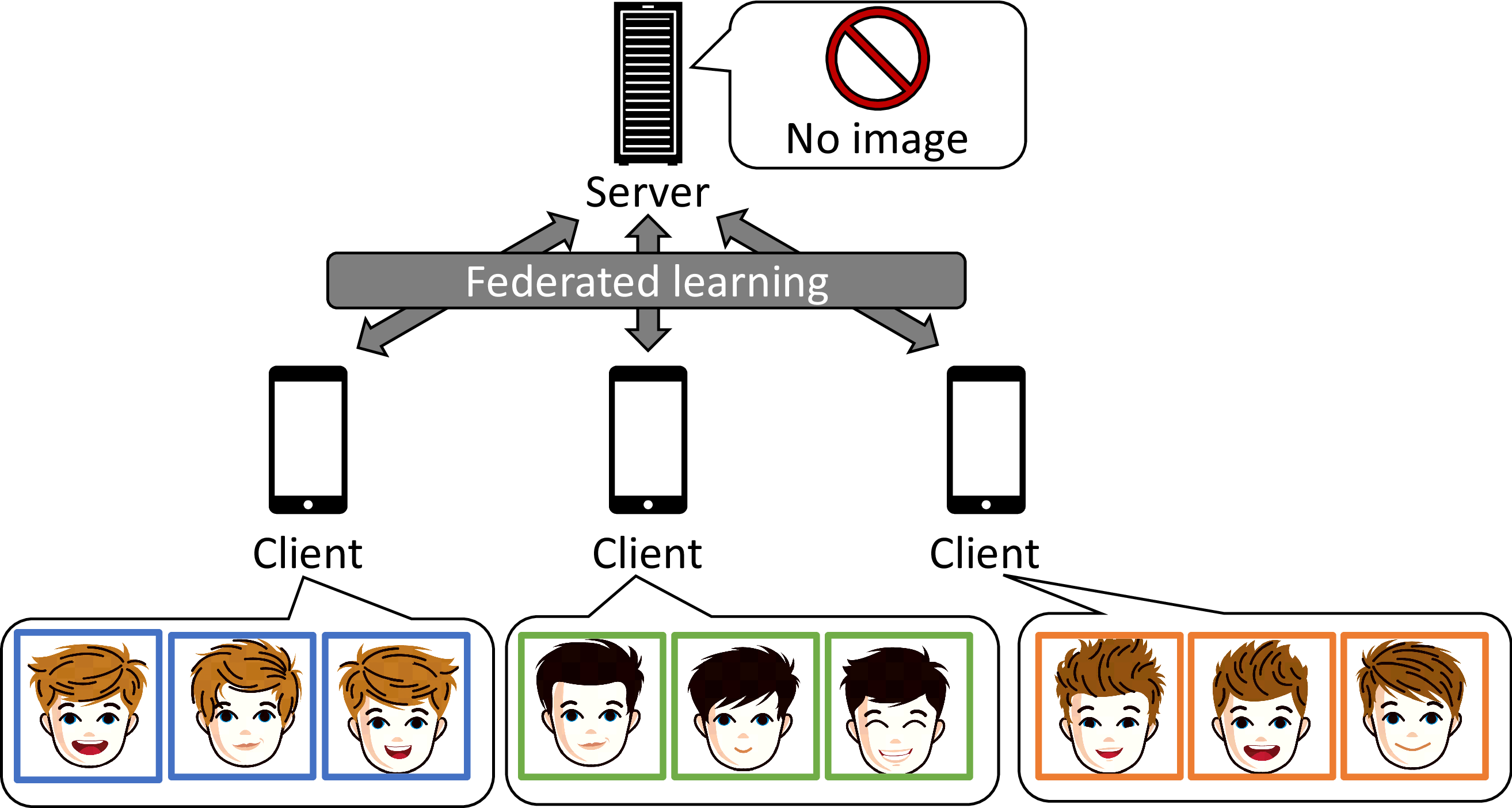}
\end{center}
\caption{Federated learning for user authentication.}
\label{fig:fl}
\end{figure}

\subsection{Federated learning}
Federated learning is a method of machine learning in which training data is not shared but distributed across multiple devices.
A typical federated learning algorithm is federated averaging  (FedAvg) \cite{mcmahan17}.
In this method, machine learning is performed while the personal data is kept in the clients, thus the privacy of training data can be preserved.

\subsection{Federated learning for user authentication}
Federated learning for user authentication is a method of learning a discriminator based on personal data such as fingerprints, faces, veins, etc., to determine a person's identity.
An overview of federated learning for user authentication is shown in Fig.\ref{fig:fl}.
Since a client is often occupied by a single person in user authentication, it is natural for a single client to contain the personal data of only one user.
Under this assumption, it is difficult to train a model that can discriminate between others because it is not possible for the client to refer to the personal data of others.

Two main approaches have been proposed to solve this problem.
The first one is \cite{Yu20,Aggarwal21}, a method where the learning server performs the learning to increase the distance between the embeddings of the others, and the second is \cite{Hosseini20,Hosseini21}, a method where the clients randomly assign representative embeddings for each class (called class embedding).
In the following, we will give an overview of each method.

\subsubsection*{FedAwS \cite{Yu20}, FedFace \cite{Aggarwal21}}
Federated Averaging with Spreadout (FedAwS) \cite{Yu20} is the algorithm to realize federated learning for user authentication by learning to increase the distance between each other's embeddings on the learning server side.
Furthermore, Aggarwal et.al. introduce FedFace, which applies FedAwS to face recognition\cite{Aggarwal21}.

Here, we introduce the algorithm of FedFace.
The learning server contains a pre-trained model parameter $\theta_{1}$, and the $i^{th}$ client contains a local dataset $S^{i}$ and a class embedding $w_1^i$.
In federated learning, the goal is to improve accuracy by updating the model parameters $\theta_{1}$ without sharing the local dataset $S^{i}$ with other clients other than the learning server and the $i^{th}$ client. 
In the $t^{th}$ round, the $i^{th}$ client receives the model parameter $\theta_{t}$ from the learning server.
The model parameter $\theta_{t}$ and the class embedding $w_t^i$ are updated by optimizing the positive loss function $l_{pos}$ using the local data $S^{i}$.
\begin{equation}
    l_{pos}(f_{\theta_{t}}(x),i) = \max{\left(0, m - (w_{t}^{i})^{T}f_{\theta_{t}}(x)\right)^2}, \label{eq:lpos}
\end{equation}
where $m$ denotes the margin parameter, $w_t^i \in \mathbb{R}^{d}$ represents the class embedding for the $i^{th}$ client and $x \in S^{i}$ is training data, $f_{\theta_{i}} : X \rightarrow \mathbb{R}^{d}$ is a face feature extractor whose parameters are $\theta_{i}$.

The updated model parameters $\theta_{t}^{i}$ and the class embedding $w_t^i$ are sent to the learning server.
The learning server aggregates the parameters $\theta_{t}^{i}$ by taking a weighted average and generates new model parameter $theta_{t+1}$.
Then, the learning server obtains the class embedding matrix $W_t = [w_t^1, w_t^2, \ldots, w_t^C]^T$ where $C$ is the number of clients.

Finally, the learning server performs spreadout regularization that minimizes the following equation:
\begin{equation}
    {reg_{sp}(W_t)} = \sum\limits_{c\in[C]}\sum\limits_{\hat{c}\neq c}\left\{\max\left(0, v - d(w_t^c, w_t^{\hat{c}})\right)\right\}^2, \label{eq:regsp}
\end{equation}
where $v$ represents the margin parameter and $d(\cdot,\cdot)$ is the function to calculate Euclidean distance.
In this way, FedFace optimizes the model parameters $\theta_t$ and the class embedding matrix $W_t$ and obtains the model with improved accuracy.

\subsubsection*{Hosseini at.el. \cite{Hosseini20,Hosseini21}}
Next, we outline the second approach \cite{Hosseini20,Hosseini21}.
In this method, class embeddings are randomly initialized and frozen on the client side, and are not shared with the learning server in order to preserve the privacy of the training data.
The clients use the fixed class embedding to train the model, and the learning server aggregates the training results and updates the model parameter.

\subsection{Challenging of related works}
Next, we will describe the challenges of federated learning for user authentication.
The first approach \cite{Yu20,Aggarwal21} discloses the class embeddings $w_{t}^i$ that contain information about the subject identity of the training data to the learning server.
These class embeddings are representative embeddings for each subject.
By using the class embeddings, the learning server could perform reconstruction of the original training data based on the model inversion attack \cite{Ziqi19}.
Therefore, user privacy is not protected by this approach, since an attacker with access to the data held by the learning server can infer certain personal data, and the class embedding $w_{t}^i$ should be kept secret from the learning server.
On the other hand, in the second approach \cite{Hosseini20,Hosseini21}, the class embeddings are assigned independently and randomly for each client, so it does not take into account the similarity of the class e.g. race or gender.
Therefore, the performance of the model obtained in the second approach is inferior to the first approach that optimizes the class embedding.

From the above discussion, as far as we know, there is no method that can perform federated learning for user authentication with both high accuracy and privacy preservation of training data.
Satisfying these two requirements is an important problem to be solved in the field of federated learning for user authentication.


\section{IPFed}
\label{sec:ipfed}
\begin{figure*}[t]
\begin{center}
\includegraphics[clip,width=15.0cm]{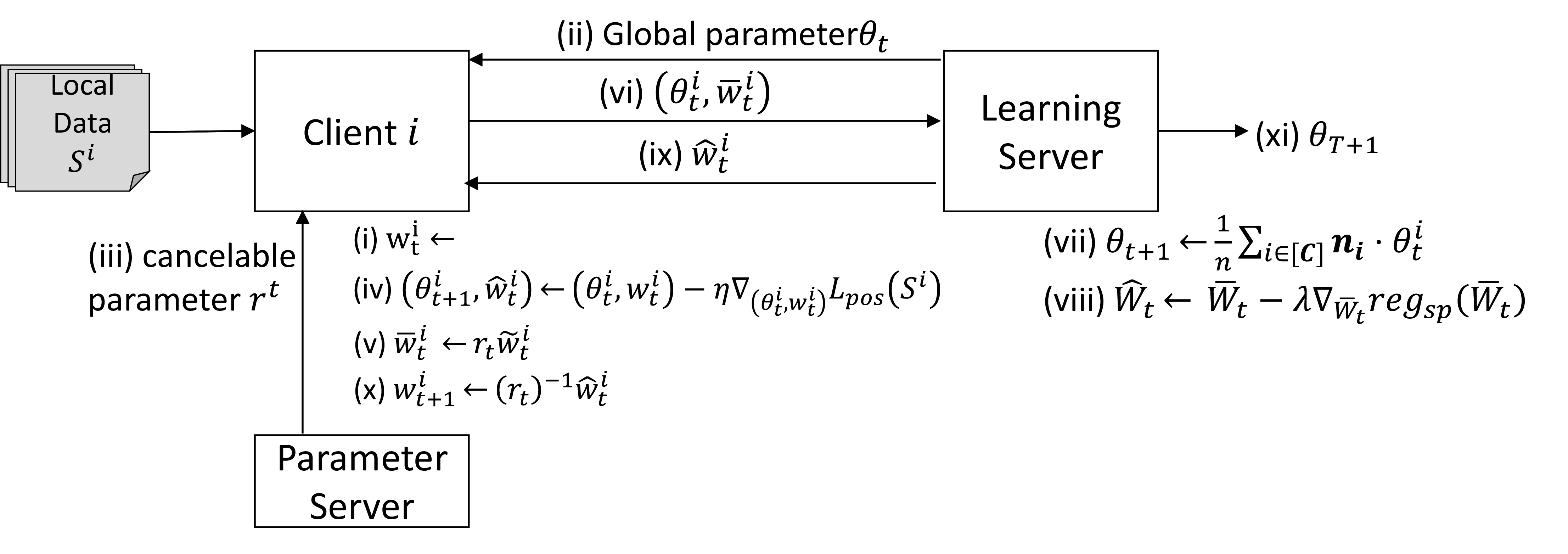}
\end{center}
\caption{The overview of IPFed.}
\label{fig:ipfed}
\end{figure*}

We propose a new method to solve the privacy and accuracy problems described in Sec.\ref{sec:related}.
We call our proposed method Identity Protected Federated Learning (IPFed).
The overview of IPFed and the training algorithm of IPFed are shown in Fig.\ref{fig:ipfed} and Algorithm \ref{alg:ipfed}, respectively.

\begin{algorithm}[!t]
    \fontsize{9pt}{9pt}\selectfont
    \SetAlgoLined
    \SetKwInOut{Input}{Input}
    \SetKwInOut{Output}{Output}

    \LinesNumberedHidden
    \Input{A pre-trained face feature extractor $f_{\theta_1}$ at the learning server and the local dataset $S^{i}$ at the $i^{th}$ client.}
    \lnlset{text}{i}
    The $i^{th}$ client initializes the class embedding $w_t^i$ by average embedding obtained by inputting $S^{i}$ into $f_{\theta_1}$.\;
    \For{$t = 1,\ldots,T$}{
        \lnlset{text}{ii}
        The learning server communicates the current global parameters $\theta_t$ to the $i^{th}$ client\;
        \lnlset{text}{iii}
        The parameter server generates random transformation parameter $r^t$ and send it to all clients\;
        \For{each of the clients $i = 1, 2, \ldots, C$}{
        \lnlset{text}{iv}
        The client updates the model parameters based on the local data $x_j^i \in S^i$\break
        $(\theta_{t}^i, \tilde{w}_{t}^{i}) \leftarrow (\theta_{t}, w_{t}^{i}) - \eta \nabla_{(\theta_{t}^i, w_{t}^{i})} L_{pos}(S^i)$\break
        where $L_{pos}(S^i) = \frac{1}{n_i}\sum\limits_{j=1}^{n_i}l_{pos}(f_{\theta_{t}^{i}}(x_j^i),i)$\;
        \lnlset{text}{v}
        The $i^{th}$ client perform random projection of class embedding $\bar{w}_{t}^{i} \leftarrow r_t \tilde{w}_{t}^{i}$\;
        \lnlset{text}{vi}
        The $i^{th}$ client sends the updated parameters $(\theta_{t}^i, \bar{w}_{t}^{i})$ back to the learning server
        }
        \lnlset{text}{vii}
        The learning server aggregates the updated parameters from all the clients\break
        $\theta_{t+1} \leftarrow \frac{1}{n}\sum\limits_{i \in [C]} n_{i} \cdot \theta_{t}^{i}$ \break
        where $n$ represents the total number of samples on all the clients\;
        \lnlset{text}{viii}
        Finally, the learning server employs the spreadout regularizer to separate the class embeddings and obtain $\hat{w}_t^i \in \hat{W}_t$\break
        $\bar{W}_{t} = [\bar{w}_{t}^1,\ldots,\bar{w}_{t}^C]$\break
        $\hat{W}_{t} \leftarrow \bar{W}_{t} - \lambda \nabla_{\bar{W}_{t}} reg_{sp}(\bar{W}_{t})$\;
        \lnlset{text}{ix}
        The learning server sends $\hat{w}_{t}^{i}$ to the $i^{th}$ client where $\hat{W}_t = [\hat{w}_t^i]$ \;
        \lnlset{text}{x}
        The $i^{th}$ client decodes class embedding and stores it: $w_{t+1}^{i} \leftarrow (r_t)^{-1} \hat{w}_{t}^{i}$
    }
    \lnlset{text}{xi}
    \Output{Output $\theta_{T+1}$}
    
    \caption{Training algorithm of IPFed.}
    \label{alg:ipfed}
    \normalsize
\end{algorithm}

IPFed consists of three entities: clients, a learning server, and a parameter server.
The parameter server is managed separately from the learning server, and we assume that it is difficult to steal data in the parameter server and the learning server at the same time.
In the following, we explain the steps of IPFed.
In the IPFed, a class embedding $w_i^t$ is initialized and stored on each client (step (i)).
The learning server sends the global parameter $\theta_t$ to each client (step (ii)).
This parameter server randomly generates a transformation parameter $r^t \in \mathbb{R}^{d \times d}$ at each round of the federated learning and sends it to each client (step (iii)), where $d$ is the dimension of the embeddings.
Each client $i$ updates the global parameter $\theta^{i}_{t}$ and the class embedding $\hat{w}^{i}_{t+1}$ (step (iv).
In addition, the client calculates the inner product of the class embedding $\tilde{w}_{t}^i$ and the transformation parameter $r^t$ (step (v)), and sends it to the learning server (step (vi)).
The learning server aggregates the global parameters $\theta^{i}_{t}$ to $\theta_{t+1}$ (step (vii)).
The learning server optimizes the received class embeddings $\bar{W}_{t}$ with spreadout similar to FedFace (step (viii)), and sends the result back to each client (step (ix)).
The client transforms the received class embedding $\hat{w}_t^i$ by the inverse matrix of the transformation parameter $(r^t)^{-1}$ (step (x)), and keeps it as its own class embedding.
After $T$ times rounds, the trained global parameter $\theta_{T+1}$ is output.

In IPFed, the class embedding is multiplied by a random transformation parameter which is secret to the learning server.
Furthermore, the updated class embedding is returned to the original feature space using the inverse matrix of the transformation parameters.
This makes it possible to perform the optimization while keeping the class embeddings secret from any server.
In the following, we show in Section \ref{subsec:derivation} that our method can perform the same optimization as FedFace even when the class embedding is kept secret, and we also show in Section \ref{subsec:privacy} that it is difficult for an attacker on any entity to obtain the user's personal data.

\subsection{Derivation of IPFed}
\label{subsec:derivation}
In this section, we analyze how the introduction of transformation parameters changes the spreadout process on the server side, and derive the conditions for learning equivalent to FedFace even with class embedding tramsformation.
First, the loss function of spreadout $reg_{sp}$ in FedFace is formulated in Eq.(\ref{eq:regsp})\cite{Aggarwal21}.
Then, the update equation using this loss function is as follows.
{\fontsize{9pt}{9pt}\selectfont
\begin{eqnarray}
& \hat{w}_{t}^c = \tilde{w}_{t}^c - \lambda \frac{\partial}{\partial \tilde{w}_{t}^c} reg_{sp} (\tilde{W}_{t}) \label{eq:wct1} \nonumber \\ 
& = \tilde{w}_{t}^c -\lambda \frac{\partial}{\partial \tilde{w}_{t}^{c}} \left\{ \sum_{c \in [C]} \sum_{\hat{c} \neq c} \left(\max \left\{0,v-d(\tilde{w}_{t}^c,\tilde{w}_{t}^{\hat{c}})\right\} \right)^2 \right\} \nonumber \\ 
& = \tilde{w}_{t}^c -\lambda \frac{\partial}{\partial \tilde{w}_{t}^c} \left\{ \sum_{\hat{c} \in [C]} 2 \left( \max \left\{0,v-d(\tilde{w}_{t}^c ,\tilde{w}_{t}^{\hat{c}}) \right\} \right)^2 \right\} \nonumber \\ 
& = \tilde{w}_{t}^c - \lambda \sum_{\hat{c} \in [C]} 4(\tilde{w}_{t}^c-\tilde{w}_{t}^{\hat{c}}) \cdot \min \left\{0,1 - \frac{v}{d(\tilde{w}_{t}^c,\tilde{w}_{t}^{\hat{c}})} \right\} \label{eq:fedface}
\end{eqnarray}}
On the other hand, in Eq.(\ref{eq:fedface}) of the IPFed, the inner product of the transformation parameter and the class embedding is input to spreadout, so if we replace $\tilde{w}_{t}^c$ with $r_t \tilde{w}_{t}^c$, we get the following equation.
{\fontsize{8pt}{8pt}\selectfont
\begin{eqnarray}
\hat{w}_t^c =& r_t \tilde{w}_{t}^c-\lambda \sum\limits_{\hat{c} \in [C]} 4(r_t \tilde{w}_t^c - r_t \tilde{w}_t^{\hat{c}}) \cdot \min \lbrace 0, 1 - \frac{v}{d(r_t \tilde{w}_t^c, r_t \tilde{w}_t^{\hat{c}}) } \rbrace \nonumber \\
\hat{w}_t^c =& r_t \left\{ \tilde{w}_{t}^c-\lambda \sum\limits_{\hat{c} \in [C]} 4(\tilde{w}_{t}^c-\tilde{w}_{t}^{\hat{c}})  \right. \left. \min \lbrace 0, 1 - \frac{v}{||r_t (\tilde{w}_{t}^c-\tilde{w}_{t}^{\hat{c}})||^2 } \rbrace \right\} \label{eq:ipfed}
\end{eqnarray}}
In addition, the learning server sends the obtained class embedding to the client, and the client transforms it into the original feature space by taking the inner product with the inverse matrix of the transformation parameters $(r_t)^{-1}$. After adding this process, Eq.(\ref{eq:ipfed}) can be expressed as follows.
{\fontsize{7.5pt}{7.5pt}\selectfont
\begin{eqnarray}
(r_t)^{-1} r_t \left\{ \tilde{w}_{t}^c - \lambda \sum_{\hat{c} \in [C]} 4(\tilde{w}_{t}^c-\tilde{w}_{t}^{\hat{c}}) \cdot \min \left\{ 0, 1 - \frac{v}{||r_t (\tilde{w}_{t}^c-\tilde{w}_{t}^{\hat{c}})||^2 } \right\} \right\}\label{eq:ipfed2}
\end{eqnarray}}
Here, if the update equation of FedFace (Eq.(\ref{eq:fedface})) and the update equation of IPFed (Eq.(\ref{eq:ipfed2})) are equivalent, then IPFed can perform equivalent learning to FedFace even if the class embeddings are kept secret.
The condition that the update equations of IPFed and FedFace are equivalent is expressed by the following equation.
{\fontsize{9pt}{9pt}\selectfont
\begin{eqnarray}
& (r_t)^{-1} r_t=I \label{eq:rt} \\
& ||r_t (\tilde{w}_{t}^c-\tilde{w}_{t}^c)||^2 = ||(\tilde{w}_{t}^c-\tilde{w}_{t}^{\hat{c}})||^2 \label{eq:rtw}
\end{eqnarray}}
where $I$ is an identity matrix.
Eq.(\ref{eq:rt}) is valid if $r_t$ is a regular matrix.
Transforming Eq.(\ref{eq:rtw}), we obtain
{\fontsize{8pt}{8pt}\selectfont
\begin{eqnarray}
(r_t (\tilde{w}_{t}^c-\tilde{w}_{t}^{\hat{c}}))^T (r_t (\tilde{w}_{t}^c-\tilde{w}_{t}^{\hat{c}})) &=&(\tilde{w}_{t}^c-\tilde{w}_{t}^{\hat{c}})^T (\tilde{w}_{t}^c-\tilde{w}_{t}^{\hat{c}}) \nonumber \\
(\tilde{w}_{t}^c-\tilde{w}_{t}^{\hat{c}})^T (r_t )^T r_t (\tilde{w}_{t}^c-\tilde{w}_{t}^{\hat{c}}) &=&(\tilde{w}_{t}^c-\tilde{w}_{t}^{\hat{c}})^T (\tilde{w}_{t}^c-\tilde{w}_{t}^{\hat{c}}) \nonumber \\
(r_t )^T r_t &=& I \label{eq:rtT}
\end{eqnarray}}
Therefore, Eq.(\ref{eq:rtT}) is valid if $r_t$ is an orthonormal matrix.
From the above, IPFed is equivalent to FedFace in terms of spreadout if the transformation parameter $r_t$ is an orthonormal matrix.

\subsection{Privacy analysis of IPFed}
\label{subsec:privacy}
In this section, we discuss how privacy protection is achieved by our IPFed.
We assume that an attacker against IPFed can obtain the data stored or communicated on any one of the three entities.
We also assume a semi-honest model in which each entity follows correct protocols.
We also assume that the goal of the attack is to obtain personal data of a specific individual.

{\bf Attacker against a client:}
From $c^{th}$client, the attacker can obtain the personal data $S^c$, the global parameter $\theta_t$, the transformation parameter $r_t$, and the class embedding $w_t^c$,$\tilde{w}_{t}^c$. 
The attacker's goal is to obtain the personal data $S^i$ (but $i \neq c$) stored on other clients from these data .
$S^c$ and $r_t$ are completely independent of the personal data $S^i$ ($i \neq c$) and do not contribute to the inference of the personal data.
The attack to estimate training data from the global parameter $\theta_t$ is called Model Inversion Attack (MIA).
However, in order to recover the training data using MIA, auxiliary information such as class labels and class embeddings are required \cite{Ziqi19}.
Since they are kept secret in the IPFed, it is difficult to perform MIA.
In addition, the class embedding ${w}_{t+1}^c$ is obtained by updating $w_t^c$ according to the distance from the class embeddings of other classes, and some information about other class embeddings may be leaked from the class embedding.
However, in federated learning, \cite{Huang21} suggests that it is difficult enough to recover the original data from the result of aggregating many gradients, and it is difficult to estimate other class embeddings from one class embedding.

{\bf Attacker against the parameter server:}
From the parameter server, the attacker can obtain the transformation parameter $r_t$. 
Since the transformation parameter is randomly generated data independent of the personal data, it is difficult to obtain any personal data.

{\bf Attacker against the learning server:}
From the learning server, the attacker can obtain the global parameters $\theta_t$,$\theta_{t+1}$, the updated global parameters $\theta_{t+1}^i$, the transformed class embeddings $\bar{w}_{t}^i$ can be obtained.
The attacker's goal is to obtain specific personal data $S^i$ from these data.
Estimation of personal data from the global parameters $\theta^t$ and $\theta_{t+1}$ is possible by gradient inversion attack, but it is extremely difficult on practical settings \cite{Huang21}.
Moreover, it is known that an attack to estimate training data from the updated global parameter $\theta_{t}^i$ can be made difficult by using secure aggregation \cite{Bonawitz17} to calculate $\theta_{t+1}$ while keeping $\theta_{t}^i$ secret.
By using these methods, user privacy can be protected under our assumption.
On the other hand, random projection is used for template protection, called cancelable biometrics \cite{Ratha01}.
This random projection is known to make it difficult to estimate the original data from the transformed data when the transformation parameter is unknown.
In general, it is known that the original data can be estimated when multiple random projections are available for the same data, but in federated learning for user authentication, the class embedding is updated and different each round, making the attack difficult.
Furthermore, by setting up a shuffle server between the learning server and the client and hiding the client's ID from the learning server, it is possible to make it difficult to assign class embeddings in different rounds.
This shuffle server can increase security.
From the above discussion, it is difficult for the learning server to infer the class embedding $\tilde{w}_t^i$ from $\bar{w}_t^i$.

In conclusion, we have confirmed that IPFed can strongly protect the privacy of training data under the assumptions defined in this paper.


\begin{table*}[!ht]
    \centering
    \vspace{-0.5\baselineskip}
    \small
    \caption{Face verification performance on standard face recognition benchmarks LFW, IJB-A and IJB-C.}
    \label{tbl:performance}
    \begin{tabular}{|l|l|l|l|l|l|}
    \hline
        \multirow{2}{*}{Method} & \multirow{2}{*}{Training data} & \multirow{2}{*}{Class embedding} & LFW & IJB-A & IJB-C \\
        ~ & ~ & ~ & Accuracy (\%) & TAR @ FAR=0.1\% & TAR @ FAR=0.1\% \\ \hline
        Baseline & Centrally aggregated & - & 99.15\% & 74.21\% & 80.60\% \\ \hline
        Fine-tuning & Centrally aggregated & Not protected & 99.30\% & 77.10\% & 82.71\% \\ \hline
        FCE & Distributed & Protected & 99.17\% & 72.71\% & 80.06\% \\ \hline
        FedFace & Distributed & Not protected & 99.22\% & 74.76\% & 79.77\% \\ \hline
        IPFed & Distributed & Protected & 99.22\% & 75.70\% & 80.73\% \\ \hline
    \end{tabular}
\end{table*}

\begin{table}[!ht]
    \centering
    \vspace{-1.0\baselineskip}
    \small
    \caption{IJB-A TAR @FAR=0.1\% on num. of subjects.}
    \label{tbl:accuracy}
    \begin{tabular}{|l|l|l|l|l|}
    \hline
        \multirow{2}{*}{Method} & \multicolumn{4}{c|}{Num. of subjects} \\ \cline{2-5}
        ~ & 50 & 100 & 500 & 1000 \\ \hline
        Fine-tuning & 75.15\% & 75.15\% & 75.76\% & 77.10\% \\ \hline
        FCE & 72.54\% & 72.21\% & 74.38\% & 72.71\% \\ \hline
        IPFed & 73.60\% & 72.82\% & 76.65\% & 75.70\%\\ \hline
    \end{tabular}
\end{table}

\subsection{Efficiency analysis of IPFed}
\label{subsec:efficiency}
In this section, we discuss the efficiency of IPFed.
In Fig.2, steps (4), (7), (11) are newly introduced steps in IPFed.
Step (4) is related to communication efficiency.
A transformation parameter $r^t \in \mathbb{R}^{d \times d}$ is very small data compared to the global parameter $\theta_t$, and its transmission has a negligible effect on the learning efficiency.
Furthermore, steps (7), (11) are related to computational efficiency on the client.
In these steps, the inner product of the matrix is computed only once, thus it has a negligible impact on learning efficiency compared to updating the global parameter in step (6).

Note that the parameter server is a newly introduced and its operating costs are newly incurred. However, since the role of the parameter server is only to generate and send the transformation parameter, there is no problem even if the server has very little computing power.

From the above discussion, it can be said that the efficiency of IPFed is comparable to that of the conventional method \cite{Aggarwal21}. However, quantitative evaluation of the efficiency of IPFed is a future work.

\section{Experiments}
\label{sec:experiments}
In this chapter, we show the effectiveness of the proposed method through experiments on face image datasets.

\subsection{Setting}
{\bf Datasets:}
We follow the setting in \cite{Aggarwal21} and used CASIA-WebFace \cite{Yi14} for training. CASIA-WebFace consists of 494,414 images of 10,575 subjects. We randomly select 9,000 subjects for pre-training and 1,000 subjects for federated learning.
To evaluate the performance of face verification, we use three datasets: LFW \cite{Huang08}, IJB-A \cite{Klare15} and IJB-C \cite{Maze18}.

{\bf Implementation:}
We use CosFace \cite{Wang18} for the face feature extractor.
Only CosFace was used as the face feature extractor according to \cite{Aggarwal21}, but evaluation using more recent multiple the face feature extractors is a subject for future work.
We use the scale parameter of 30 and the margin parameter of 10 for the CosFace loss function.
The margin parameters are $m = 0.9$ and $v = 0.7$.
The parameter $\lambda$ = 25 in Eq.(\ref{eq:wct1}).
We perform federated learning with a communication round of 200 and a learning rate of 0.1.

\subsection{Evaluation}
First, the face verification performance of each method for 1000 clients is shown in the Table.\ref{tbl:performance}.
The comparison methods are shown below:

{\bf Baseline} A typical CosFace model, pre-trained on 9000 subjects.

{\bf Fine-tuning} A fine-tuned model of Baseline using CosFace on 1000 subjects.

{\bf FedFace} A model trained according to \cite{Aggarwal21}.

{\bf IPFed} A model trained based on the proposed random projection approach.

{\bf Fixed class embedding (FCE)} A model trained by fixing the class embedding to the initial data.
In FCE, the class embedding does not need to be shared with the server, and secure federated learning can be performed.

As shown in Table.\ref{tbl:performance}, IPFed achieves the same level of accuracy as FedFace.
This means that the random projection based spreadout in IPFed is equivalent to the spreadout in FedFace.
On the other hand, FCE is less accurate than IPFed.
This is due to the lack of class embedding optimization, which means that the spreadout in IPFed contributes to the accuracy improvement.

Furthermore, the accuracy against the number of subjects used for federated learning is shown in Table\ref{tbl:accuracy}.
The IPFed achieves higher accuracy than the FCE for any of the number of the subjects.
This indicates that sharing and optimizing class embedding improve the accuracy.
However, while Fine-tuning has a monotonic increase in accuracy for the number of the subjects, IPFed does not.
This may be due to the fact that the hyperparameters in learning are not optimal for the number of the subjects, so automatic adjustment of hyperparameters is a future task.

In addition, in our experiments, only accuracy was evaluated, and attack defense performance was only theoretically evaluated.
This is also a topic for future work.

\section{conclusion}
\label{sec:conclusion}
In this paper, we focused on the problem of personal data leakage from class embedding in federated learning for user authentication, and proposed IPFed, which performs federated learning while protecting class embedding.
We proved that IPFed, which is based on random projection for class embedding, can perform learning equivalent to the state-of-the-art method.
We evaluate the proposed method on face image datasets and confirm that the accuracy of IPFed is equivalent to that of the state-of-the-art method.
IPFed can improve the model for user authentication while preserving the privacy of the training data.

\printbibliography

\end{document}